\pdfoutput=1

\documentclass[11pt]{article}

\usepackage[]{acl}

\usepackage{times}
\usepackage{latexsym}
\usepackage{amssymb}
\usepackage[linesnumbered,ruled,vlined]{algorithm2e}

\usepackage[T1]{fontenc}

\usepackage[utf8]{inputenc}

\usepackage{microtype}
\usepackage{tabularx}
\usepackage{url}

\usepackage{booktabs}
\usepackage{multirow}
\usepackage{graphicx}
\usepackage{amsthm}
\usepackage{amsmath}

\theoremstyle{definition}
\newtheorem{exmp}{Example}[section]

\newcommand{\ACRO}[1]{\textsc{#1}}
\newcommand{\ARRAU}{\ACRO{arrau}}
\newcommand{\BERT}{\ACRO{bert}}
\newcommand{\CONLL}{\ACRO{conll}}
\newcommand{\CRAC}{\ACRO{crac}}
\newcommand{\FFNN}{\ACRO{ffnn}}
\newcommand{\LEA}{\ACRO{lea}}
\newcommand{\JOINT}{\ACRO{joint}}
\newcommand{\PRETRAIN}{\ACRO{pre-trained}}
\title{Stay Together: A System for Single and Split-antecedent Anaphora Resolution}

\author{Juntao Yu$^1$, Nafise Sadat Moosavi$^2$, Silviu Paun$^1$, Massimo Poesio$^1$ \\
  $^1$Queen Mary University of London \\
  $^2$UKP Lab, Technische Universität Darmstadt\\
  $^1${\tt \{juntao.yu, s.paun, m.poesio\}@qmul.ac.uk} \\
  $^2${\tt moosavi@ukp.informatik.tu-darmstadt.de}}

\begin{document}
\maketitle
\begin{abstract}
The state-of-the-art  on basic, single-antecedent anaphora has greatly improved  in recent years. 
Researchers have therefore started to pay more attention to 
more complex cases of anaphora such as
split-antecedent anaphora, as in \textit{\underline{Time-Warner} is considering a legal challenge to \underline{Telecommunications Inc's} plan to buy half of Showtime Networks Inc--a move that could lead to all-out war between \textbf{the two powerful companies}.} 
Split-antecedent anaphora is rarer and more complex to resolve than single-antecedent anaphora; 
as a result, it is not annotated in many datasets designed to test coreference, and
previous work on resolving this type of anaphora was carried out in unrealistic conditions that assume gold mentions and/or gold split-antecedent anaphors are available. 
These systems also focus on split-antecedent anaphors only.
In this work, we introduce a system that resolves both single and split-antecedent anaphors, and evaluate it in a more realistic setting that uses predicted mentions. We also start addressing the question of how to evaluate single and split-antecedent anaphors together using standard coreference evaluation metrics.\footnote{The code is available at \url{https://github.com/juntaoy/dali-full-anaphora}
}
\end{abstract}

\section{Introduction}
Thanks in part to the  latest developments in deep neural network 
architectures 
and 
contextual word embeddings (e.g., ELMo \cite{peters2018elmo} and {\BERT} \cite{devlin2019bert}),
the performance of models for single-antecedent anaphora resolution has greatly improved \cite{wiseman2016learning,clark2016improving,lee2017end,lee2018higher,kantor-globerson-2019-coreference,joshi2019spanbert}. 
So recently, the attention has turned to 
more complex 
cases of anaphora, 
such as 
anaphora requiring some sort of commonsense knowledge as in the Winograd Schema Challenge \cite{rahman&ng:EMNLP2012,peng-et-al:NAACL2015,liu-ijcal-17-winograd,Sakaguchi-aaai-20-winograde};
pronominal anaphors that cannot be resolved purely using gender \cite{webster-et-al:TACL2018},
bridging reference \cite{hou-2020-acl,yu-etal-2020-bridging},
discourse deixis \cite{kolhatkar-hirst-2014-resolving,marasovic-etal-2017-mention,kolhatkar-et-al:CL18} 
and, finally, split-antecedent anaphora  \cite{zhou&choi:COLING2018,yu-etal-2020-plural} - plural anaphoric reference in which the two antecedents are not part of a single noun phrase. 

However, 
a number of hurdles have to be tackled  when trying to study these cases of anaphora, ranging from the lack of annotated resources to the rarity of some of these phenomena in the existing ones. 
Thus, most previous work on resolving these anaphoric relations focused on developing dedicated systems for the specific task. 
The systems are usually enhanced by transfer-learning to utilise extra resources, as those anaphoric relations are  sparsely annotated. The most frequently used extra resource is single-antecedent anaphors. Due to the complexity of these tasks,  previous work is usually based on assuming that either gold anaphors \cite{hou-2020-acl,yu-etal-2020-plural} or gold mentions  \cite{zhou&choi:COLING2018,yu-etal-2020-bridging} are provided. 
By contrast, in this work we introduce a system that resolves both single and split-antecedent anaphors, and is evaluated in a more realistic setting that does not rely on  gold anaphors/mentions. 
We evaluate our system on the {\ARRAU}  corpus \cite{poesio_anaphoric_2008,uryupina-et-al:NLEJ}, in which both single and split-antecedent anaphors are annotated, although the latter are much rarer than the former. 
We use the state-of-the-art coreference resolution system on {\ARRAU}  \cite{yu-etal-2020-cluster} as our base system for single-antecedent anaphors. 
This cluster-ranking system 
interprets  single-antecedent anaphors, singletons and non-referring expressions jointly. 
In this work, we extend the system to resolve  split-antecedent anaphors. 
The extended part of the system shares  mention representations and candidate clusters with the base system, and outputs binary decisions between a mention and individual candidate clusters. We configure our system to learn the split-antecedent part and the base system in both {\JOINT} and {\PRETRAIN} fashion. 
The results show both versions work much better than  naive baselines based on heuristics and random selection.  
The {\PRETRAIN} version 
works equally well as 
the {\JOINT} version on split-antecedent anaphors,
but it is better for 
the other aspects of anaphoric interpretation.

In the paper we also begin to address the question of how  a system carrying out both single and split-antecedent anaphora resolution should be evaluated. 
Specifically, we introduce an extended version of {\LEA} \cite{moosavi-strube-2016-coreference}, a standard coreference metric which can be used to give partial credit for resolution,  to evaluate single and split-antecedent anaphors together. 
Using this metric, we find that our best model achieves a better {\LEA} score than the baselines. 

We further evaluate our best system in the gold setting to compare with the \newcite{yu-etal-2020-plural} system. The model achieved better performance when compared to their system that is designed solely for split-antecedent task.

\section{Related Work}
\subsection{Neural Approaches to Single-antecedent Anaphora Resolution}
Single-antecedent anaphora resolution is an active research topic. 
The first neural model was introduced by \newcite{wiseman2015learning}
and 
later extended in \cite{wiseman2016learning}.
\newcite{clark2016improving} introduced a hybrid cluster/mention-ranking approach, 
whereas \newcite{clark2016deep} adapted reinforcement learning to a mention-ranking model. 
%
\newcite{lee2017end} introduced the first end-to-end system, performing mention detection and coreference resolution jointly. 
The \newcite{lee2017end} system was also simpler than  previous systems, using only a small number of hand-coded features. 
As a result, the  \newcite{lee2017end} system has become the blueprint for most subsequent systems. \newcite{lee2018higher} and \newcite{kantor-globerson-2019-coreference} showed that employing contextual ELMo \cite{peters2018elmo} and {\BERT} \cite{devlin2019bert} embeddings in the system by \newcite{lee2017end}  can significantly improve  performance. 
\cite{joshi2019bert,joshi2019spanbert} fine-tuned  {\BERT} and SpanBERT to further improve performance. 
Recently, \newcite{wu-etal-2020-corefqa} framed coreference resolution task as question answering and showed that the additional pre-training on a large question answering dataset can further improve  performance. 
However, those systems are only focused on  single-antecedent anaphors and do not consider the other  anaphoric relations. 
\vspace{-5pt}

\subsection{Other Aspects of Anaphoric Interpretation}

Interpreting nominal expressions with respect to a discourse model is not simply a matter of identifying identity links; 
it also involves recognizing that certain potential anaphors are in fact non-referring, or singletons;
other expressions refer to entities which have to be introduced in the discourse model via accomodation processes involving for instance the construction of a plural object out of other entities, as in the case of split-antecedent anaphors;
other expressions again are related to existing entities by associative relations, as in \textit{one}-anaphora or bridging reference.
These other anaphoric interpretation processes are much less studied, primarily because the relevant information is not annotated in the dominant corpus for coreference, OntoNotes \cite{pradhan2012conllst}.
Systems such as  the Stanford Deterministic Coreference Resolver \cite{lee-et-al:CL13} 
do use linguistically-based  heuristic rules to recognize and filter singletons and non-referring expressions,
but these aspects of the system are not evaluated.
Carrying out such an evaluation requires a corpus with richer anaphoric annotations, such as {\ARRAU} \cite{uryupina-et-al:NLEJ}. 

\newcite{yu-etal-2020-cluster} is the only neural system that targets singletons and non-referring expressions. The system uses the mention representation from  \newcite{lee2018higher,kantor-globerson-2019-coreference} and applies a cluster-ranking algorithm to incrementally attach mentions directly to their clusters. 
\newcite{yu-etal-2020-cluster} showed that  performance on single-antecedent anaphors improves  by up to 1.4 p.p. when jointly training the model with non-referring expressions and singletons. We use  \newcite{yu-etal-2020-cluster} as our base system, and extend it to resolve split-antecedent anaphors. 

A few systems  resolving split-antecedent anaphors have been proposed in recent years. \newcite{vala-etal-2016-antecedents} introduced a 
system to resolve  plural pronouns \textit{they} and \textit{them} in a fiction corpus they themselves annotated. 
\newcite{zhou&choi:COLING2018} introduced an entity-linking corpus based on the transcripts of the \textit{Friends} sitcom. 
The mentions (including plural mentions) are annotated if they are linked to the main characters. 
Coreference clusters are then created for mentions linked to the same entities.
One issue with this corpus is that it is 
mainly created for entity-linking, so it is problematic as a coreference dataset, as many mentions are linked to general entities that are not annotated in the text. 
\newcite{zhou&choi:COLING2018} trained a  \ACRO{cnn} classifier to determine the relation between mention pairs, 
jointly performing single and split-antecedent resolution. 

Another issue with this work is evaluation. 
\newcite{zhou&choi:COLING2018} evaluate their system using the standard {\CONLL} scorer; in order to do this, they encode split-antecedent anaphora by adding the plural mention to each cluster. 
So, for instance, in \textit{John met Mary. They went to the movies}, they would have two  gold clusters: \{John, They\} and \{Mary, They\}.
This is clearly problematic, as \textit{They} is not a mention of the individual entity John, but of the set consisting of John and Mary.
In this work, we 
propose an alternative,
an extended version of {\LEA} \cite{moosavi-strube-2016-coreference} that does joint evaluation of single/split-antecedent anaphors by explicitly representing plural entities.  

\newcite{yu-etal-2020-plural} introduced the first system to resolve all split-antecedent anaphors annotated in the {\ARRAU} corpus. Their work focuses on the data sparsity problem;  split-antecedent anaphora resolution is helped using   four auxiliary corpora created from a crowdsourced corpus and other anaphoric annotations in the {\ARRAU} corpus. 
However, their approach focuses on split-antecedent anaphora only, and assumes  gold split-antecedent anaphors and gold mentions are provided during the evaluation, which is not realistic. In this work, we resolve both single and split-antecedent anaphora and evaluate our system on predicted mentions.

\section{The Resolution Method}

\subsection{The Base System}
In this work, we use the system of \newcite{yu-etal-2020-cluster} as starting point, and  extend it  to handle split-antecedent anaphora.  \newcite{yu-etal-2020-cluster}  is a cluster-ranking system that jointly processes single-antecedent anaphors, singletons and non-referring expressions. 
The system uses the same mention representations as in \newcite{lee2018higher,kantor-globerson-2019-coreference}. 
The input to the system is a concatenation of  contextual {\BERT} \cite{devlin2019bert} embeddings,  context-independent \ACRO{glove} embeddings \cite{pennington2014glove} and  learned character-level embeddings based on convolutional neural network (\ACRO{cnn}s). The system then uses a multi-layer 
\ACRO{bilstm} 
to encode the document at the sentence level to create the word representations ($T_i$). The candidate mention representations ($M_i$) are created by the concatenation of the word representations at the start/end positions of the mention as well as a weighted sum of all the tokens within the mention boundary. After that, the candidate mentions are pruned according to their mention scores ($s_m(i)$) computed by applying a feed-forward neural network ({\FFNN}) to the $M_i$. The top-ranked candidate mentions are then used by the cluster-ranking model to form the entity clusters and to identify the non-referring expressions.

The cluster-ranking model incrementally  links the candidate mentions  to the clusters  according to the scoring function ($s(i,j)$) between candidate mention $M_i$ and partial clusters created so far ($C_{i-1}^j$). More precisely, $s(i,j)$ is defined as:

\vspace{-10pt}
\small
$$
s(i,j) = \left\{
  \begin{tabular}{ll}
  $s_{no}(i)$& $j = \textsc{no}$ \\
  $s_{nr}(i) + s_m(i)$&  $j = \textsc{nr}$\\
  $s_{dn}(i) + s_m(i)$ & $j = \textsc{dn}$\\
  $s_m(i) + s_c(j)+ s_{mc}(i,j)$& $j \in C_{i-1}$
  \end{tabular} 
  \right.
$$
\normalsize
where $s_{no}(i)$, $s_{nr}(i)$ and $s_{dn}(i)$ are the likelihood for a candidate mention to be a non-mention (\textsc{no}), 
a non-referring expression (\textsc{nr}) or a discourse new mention (\textsc{dn}) respectively. 
$s_m(i)$, $s_c(j)$ and $s_{mc}(i,j)$ are the mention scores (computed for mention pruning), cluster scores (a weighted sum of $s_m$ for the mentions in the cluster) and cluster-mention pairwise scores. 
The system 
employs additional methods 
to enhance 
performance--e.g., keeping cluster histories and training the system on the oracle clusters. We refer the reader to \cite{yu-etal-2020-cluster} for more details. We use the default settings of the system in our experiments.

\subsection{Resolving Split-antecedent Anaphors}

To resolve  split-antecedent anaphors, we follow \newcite{yu-etal-2020-plural} who framed the task as a binary classification task. The system uses a scoring function to assign each cluster-mention pair a score $s_p(i,j)$ 
specifying 
the likelihood 
that that cluster 
is one 
of the split-antecedents of the mention. During training, we add a dummy score ($s_\epsilon(i) = 0$) 
for 
the cases in which 
a mention is not a split-antecedent anaphor. 
Formally, $s_p(i,j)$ is calculated as follows:

\vspace{-10pt}
\small
$$
s_p(i,j) = \left\{
  \begin{tabular}{ll}
  $0$& $j = \epsilon$ \\
  $s_m(i) + s_c(j)+ s_{pmc}(i,j)$& $j \in C_{i-1}$
  \end{tabular} 
  \right.
$$
\normalsize
The extension for split-antecedents 
uses the same mention/cluster representations as well as the candidate mentions/clusters 
of the single-antecedent component.
This benefits the split-antecedent anaphors part of the system, that can share the representations learned from more numerous single-antecedent anaphors. 
As a result, the extension shares the same $s_m(i)$ and $s_c(j)$ scores as the base system. 
$s_{pmc}$ is calculated by applying a {\FFNN} to the cluster-mention pairwise representations. At test time, we convert $s_p(i,j)$ into probabilities ($p_p(i,j)$), and assign split-antecedents to plural mentions when the $p_p(i,j)$ between the plural mentions and the candidate clusters are above the threshold (e.g., $0.5$).  $p_p(i,j)$ is calculated by applying a sigmoid function to $s_p(i,j)$:

\vspace{-10pt}
$$p_p(i,j) = \frac{1}{1+e^{-s_p(i,j)}} $$

To make sure the final system outputs (single-antecedent anaphors, singletons, non-referring expressions and split-antecedent anaphors) do not contradict  each other, we  only allow discourse-new mentions to become split-antecedent anaphors. We also constrain  split-antecedent anaphors to have at least two and at most five antecedents. 

Since we are working with  predicted clusters, to evaluate using  lenient and strict scores as in \newcite{yu-etal-2020-plural}, we need to find a way to align  the predicted clusters with the gold clusters. Here we use the standard coreference 
alignment function 
CEAF$_{\phi4}$ to align predicted and gold clusters. The alignment between predicted and gold clusters is at the centre of the CEAF$_{\phi4}$ scores, which gives exactly what we need for our evaluation.

\subsection{Training Strategies}
To train, we add 
to the original loss ($loss_{s}$) a second dedicated loss ($loss_{p}$) for  split-antecedent anaphors. 
We use  marginal log-likelihood loss, and optimize on all  oracle clusters that belong to the gold split-antecedent cluster list \textsc{gold}$_p(i)$ of split-antecedent anaphors $M_i$.
Formally,

\vspace{-10pt}
$$loss_p = log \prod_{j=1}^{N}\sum_{\hat{c} \in \textsc{gold}_p(j)} s_p(\hat{c},j)$$
%
Since the vast majority of  mentions (99\%) are negative examples (non-split-antecedent anaphors),  training is highly imbalanced. 
So during training we  also use the mentions from the same cluster as the split-antecedent anaphors as additional positive examples. 
In this way we managed to nearly double
the number of positive training examples. 
We multiply the losses of the negative examples an adjustment parameter $\alpha$ to balance the training.

We train our system  both in {\JOINT} and {\PRETRAIN} mode.
For \textbf{{\JOINT}} learning, we train our system on the sum of two losses and weigh them by a $\beta$ factor that determines the relative importance of the losses. Formally, we compute the joint loss as follows:

\vspace{-10pt}
$$loss_{j} = (1 - \beta)loss_s + \beta loss_p $$
To use a joint loss the split-antecedent part of the system can have an impact on the mention representations hence might lead to better performance. 

Our \textbf{{\PRETRAIN}} approach is based on the hypothesis that mention/cluster representations trained on the single-antecedent anaphors are sufficient  as pre-trained embeddings for downstream tasks like split-antecedent anaphors. 
The {\PRETRAIN} approach  minimises the 
changes to the base system, and one can even reuse the models trained solely for the base system. The training for the split-antecedent part is inexpensive. We use the pre-trained models for our base system to supply mention/cluster representations and other necessary information and optimise the split-antecedent part of the system solely on  $loss_p$.  

\section{Evaluating Coreference Chains with Split Antecedents}
\label{section:lea}


If the interpretation of a split-antecedent anaphor were only 
given credit 
when all  antecedents are correctly detected and grouped together, 
without giving any reward to systems that find at least some of the antecedents, systems that get closer to the gold would be unfairly penalized, 
particularly for the cases with 3 or more split antecedents (25\% in our data).
Consider example~\ref{ex-1}, in which ``their$_{i,j}$'' refers to the set \{``Mary'', ``John''\}, 
and ``they$_{i,j,p}$'' to the set \{``Mary'', ``John'', ``Jane''\}.
And take two systems A and B that resolve 
``their$_{i,j}$'' to \{``Alex'', ``Jane''\} and \{``Mary'', ``Jane''\}, respectively 
and ``they$_{i,j,p}$'' to \{``Alex''\} and \{``Mary$_i$'', ``John$_j$''\}, respectively.
Neither system is perfect, 
but intuitively, system B is more accurate in resolving split-antecedent anaphors (it correctly identifies 1 antecedent of ``their$_{i,j}$'' and 2 of ``they$_{i,j,p}$'', versus 0 for A)--yet both systems will receive the same 0 score 
if only a perfect match is credited.

\begin{exmp}
	Mary$_i$ and John$_j$ were on their way to visit Alex$_k$ when Mary$_i$ saw Jane$_p$ on their$_{i,j}$ way and realized 
	they$_{i,j,p}$  all wore the same shirt.
	\label{ex-1}
	
\end{exmp}
This example indicates  that in order to score a system carrying out both single and split-antecedent resolution three issues have to be addressed. 
First of all, it is necessary to have some way to represent plural entities.  
Second, we need some way of ensuring that systems that propose different but equivalent resolutions for split-antecedent plurals score the same.
Third,  we need a metric allowing some form of partial credit.\footnote{This third issue is the reason why \cite{vala-etal-2016-antecedents,yu-etal-2020-plural} used lenient metrics for scoring split-antecedent resolution, 
although ones that did not score single antecedent resolution as well.}
We discuss how we addressed each issue in turn. 

\paragraph{Plural mentions} First of all, we propose to have two types of mentions in our coreference chains: in addition to the standard \textbf{individual} mentions (``Mary''), we also allow  \textbf{plural} mentions (\{``Mary'', ``Jane''\}).

\paragraph{Normalizing coreference chains} As Example~\ref{ex-1} shows, a  text may contain multiple individual mentions of the same entity that participate in a plural mention (e.g. `Mary'). 
Plural mentions whose antecedents are mentions of the same entity should be equivalent. 
To do this, we use the first mention of each gold coreference chains as the representative of the entity. We \textbf{normalize} every plural mention in a system- produced  coreference chain by 
(i) aligning the system-produced coreference chains for the individual mentions in the plural mention to the gold coreference chains using \ACRO{ceaf}, and 
(ii) replacing each individual mention in the plural mention with the first mention in the aligned gold coreference chains. 

\paragraph{Partial credit}
A natural way to obtain a scorer for coreference resolution 
giving partial credit is to
extend 
the {\LEA} evaluation metric \citep{moosavi-strube-2016-coreference} to handle split-antecedents.
For each entity $e$, {\LEA} evaluates 
(a) how important is $e$, and 
(b) how well it is resolved. 
Thus, for computing recall, {\LEA} evaluates a set of system-detected entities $E$ as follows:\footnote{We can compute precision by switching the role of system and key entities in {\LEA} computations.} 
\begin{equation}
	\frac{\sum_{e \in E} \text{importance}(e) * \text{resolution-score}(e)}{\sum_{e \in E} \text{importance}(e)}
\end{equation}
where \emph{resolution-score} is the ratio of correctly resolved coreference links in the entity, and the \emph{importance} measures how important is entity $e$ in the given text. In the default implementation, \emph{importance} is set to the size of the entity. However, it can be adjusted based on the use case. 

Let $e$ be an entity in the system output $E$ consisting of $n$ mentions, and $K$ be the set of gold entities.
The \emph{resolution-score} ($RS$) of $e$ is computed as:
\begin{equation}
	RS(e) = \frac{1}{|\mathbb{L}(e)|}\sum_{l \in \mathbb{L}(e)}  \mathbb{B}(l, K)
\end{equation}
where $\mathbb{L}(e)$ is the set of all coreference links in $e$\footnote{There are $\frac{n(n-1)}{2}$ coreference links in $e$.}, and $\mathbb{B}(l, K)$ is defined as
\begin{equation}
\mathbb{B}(l, K)= 
	\begin{cases}
		1 &   \{\exists_{k\in K}  | l \in \mathbb{L}(k)\}\\
		0 &  \text{otherwise}
	\end{cases}
	\label{belong-eq}
\end{equation}

(\ref{belong-eq}) states that for each coreference link $l$ in system entities, the system receives a reward of one if $l$ also exists in gold entities, and zero otherwise. If any of the mentions that are connected by $l$ is a partially resolved plural mention, the system receives a zero score.

To extend {\LEA} to handle split-antecedents, we change $\mathbb{B}$ to also reward a system if any of the corresponding mentions of $l$, i.e., mentions that are connected by $l$, is a plural mention and is partially resolved.
Let $\hat{\mathbb{P}}(m)$ be an ordered list of all subsets of $m$, including $m$, by descending order of their size.
If $m$ is a singular mention, $\hat{\mathbb{P}}$ will only contain \{$m$\}.
If $m$ is a plural mention, $\hat{\mathbb{P}}$ will contain $m$ as well as all the subsets of $m$'s containing mentions. For instance, $\hat{\mathbb{P}}$(\{``Mary'', ``John''\})=[ \{``Mary'', ``John''\}, \{``John''\}, \{``Mary''\}].
Assuming the corresponding mentions of $l$ are $m_i$ and $m_j$, we update $\mathbb{B}(l, K)$ as follows:

\small
\[  
	\begin{cases}
		\frac{|s_i|*|s_j|}{|m_i|*|m_j|} &   \{\exists_{k\in K,s_i \in \hat{\mathbb{P}}(m_i), s_j \in \hat{\mathbb{P}}(m_j)}  | l_{s_i,s_j} \in \mathbb{L}(k)\}\\
		\frac{|m_i|*|m_j|}{|m_k|*|m_p|} &   \{\exists_{k\in K,m_i \in \hat{\mathbb{P}}(m_k), m_j \in \hat{\mathbb{P}}(m_p)}  | l_{m_k,m_p} \in \mathbb{L}(k)\}\\
		0 &  \text{otherwise}
	\end{cases}
	\label{belong-plural}
\]

\normalsize
where $l_{s_i,s_j}$ is the link connecting $s_i$ and $s_j$ that are the largest subset of $\hat{\mathbb{P}}(m_i)$ and $\hat{\mathbb{P}}(m_j)$, respectively, that exist in gold entities and are coreferent.  $m_k$ and $m_p$ are gold coreferring mentions that $m_i$ and $m_j$ are a subset of, respectively. 

For instance, consider the system chain \{$m_1$=\{``Mary'', ``Jane''\}, $m_2$=``their$_{i,j}$''\} for Example~\ref{ex-1}. The coreference link between $m_1$ and $m_2$ does not exist in the gold entities. However, $m_1$ is a subset of a gold mention, i.e., $m_k$=\{``Mary'', ``John'', ``Jane''\}, and $m_1 \subset \hat{\mathbb{P}}(m_k)$.
Therefore, system B receives a reward of $\frac{2*1}{3*1}$ for resolving the coreference link between $m_1$ and $m_2$ based on $RS$.
 
\paragraph{Importance}
As discussed, the number of entities that contain split-antecedents in our annotated data is negligible compared to entities with singular mentions.
Therefore, we will not see a big difference in the overall score when the system resolves both singular and plural mentions.
In order to put more emphasize on harder coreference links, i.e., resolving split-antecedents, we adapt the \emph{importance} measure to assign a higher weight to entities containing split-antecedent as follows:

\vspace{-15pt}
\begin{equation}
	\text{importance}(e)= \frac{\text{importance-factor}(e)*|e|}{\sum_{e_i} \text{importance-factor}(e_i)*|e_i|} \nonumber
\end{equation}
The \emph{importance-factor} assigns Imp$_{split}$ times higher importance on plural entities compared to entities of singular mentions:
\small
\begin{equation}
	\text{importance-factor}(e) = 
	\begin{cases}
		\text{Imp}_{split} &  \quad \text{If } e \text{ is a plural entity} \\
		1 &  \quad \text{If } e \text{ is singular}
	\end{cases}	\nonumber
\end{equation}
\normalsize


\section{Experiments}

\subsection{Datasets}
We evaluated our system on the \ACRO{rst} portion of the {\ARRAU} corpus \cite{uryupina-et-al:NLEJ}. 
{\ARRAU}  provides a wide range of anaphoric information  (referring expressions including singletons and non-referring expressions; split-antecedent plurals; generic references; discourse deixis; and bridging references) and was used in the {\CRAC} shared task \cite{poesio2018crac}; 
\ACRO{rst} was the main evaluation subset in that task;
the \ACRO{rst} portion of the {\ARRAU} corpus consists of 1/3 of the Penn Treebank (news texts). 
Table \ref{table:crpus} summarizes the key statistics about the corpus.

\begin{table}[t]
    \centering
    \small
    \begin{tabular}{l l l}
    \toprule
     & \bf Train/Dev &\bf Test \\
    \midrule
    Documents &353&60\\
    Sentences &7524&1211\\
    Tokens &195676&33225\\
    Mentions &61671& 10341\\
    Singletons&26368&4158\\
    Non-referring expressions&8159&1391\\
    Single-antecedent anaphors&20127&3568\\
    Split-antecedent anaphors & 356&60\\
    Split-antecedents&878&137\\
    \bottomrule
    \end{tabular}
    \caption{Statistics about the corpus used for evaluation.}
    \label{table:crpus}
\end{table}
\begin{table}[t]
    \centering
    \small
    \begin{tabular}{l l}
    \toprule
    \bf Parameter & \bf Value \\
    \midrule
    BiLSTM layers/size/dropout &3/200/0.4\\
    FFNN layers/size/dropout & 2/150/0.2\\
    CNN filter widths/size& [3,4,5]/50\\
    Char/Glove/Feature embedding size&8/300/20\\
    {\BERT} embedding layer/size &Last 4/1024\\
    Embedding dropout & 0.5\\
    Max span width ($l$)&30\\
    Max num of clusters&250\\
    Mention/token ratio &0.4\\
    Optimiser & Adam (1e-3)\\
    Non-referring method&Hybrid\\
    Prefiltering threshold&0.5\\
    Adjustment parameter ($\alpha$)&0.01\\
    Loss weight ($\beta$)&0.1\\
    \bottomrule
    \end{tabular}
    \caption{Hyperparameters for our models.}
    \label{table:config}
\vspace{-15pt}
\end{table}

\subsection{Separate and Joint Evaluation Methods}

In \textbf{separate evaluation}, 
we follow  standard practice to report {\CONLL} average F1 score 
(macro average of MUC, B$^3$ and CEAF$_{\phi4}$) for single-antecedent anaphors, 
and F1 scores for non-referring expressions. 
For split-antecedent anaphors, we report three F1 scores: the strict F1 score that only gives credit when both anaphors and all their split-antecedents are resolved correctly\footnote{Here we report F1 instead of accuracy used in \newcite{yu-etal-2020-plural} as our evaluation is based on predicted mentions.};
the lenient F1 score that gives credit to anaphors that resolved partially correct \cite{vala-etal-2016-antecedents}; 
and the anaphora recognition F1 score.

For \textbf{joint evaluation} of single/split-antecedent anaphors, we report the {\LEA} score using the upgraded script described in Section \ref{section:lea}.

\subsection{Hyperparameters}
We use the default parameter settings of  \newcite{yu-etal-2020-cluster}  and use their hybrid approach for handling the non-referring expressions. 
The split-antecedent part of the system uses an {\FFNN} with two hidden layers and a hidden size of 150. The negative example loss adjustment parameter $\alpha$ and the loss weight parameter $\beta$ (used for {\JOINT} learning) are set to 0.01 and 0.1 respectively after tuning on the development set. Table \ref{table:config} 
provides details on 
our parameter settings.

\begin{table*}[t]
\centering
\resizebox{\textwidth}!{
\begin{tabular}{lccccccccccccc}
\toprule
&\bf CoNLL&\multicolumn{3}{c}{\bf Non-referring}&\multicolumn{3}{c}{\bf Anaphora Rec$_{split}$ }&\multicolumn{3}{c}{\bf Lenient$_{split}$}&\multicolumn{3}{c}{\bf Strict$_{split}$}\\
&\bf F1&\bf R&\bf P&\bf F1&\bf R&\bf P&\bf F1&\bf R&\bf P&\bf F1&\bf R&\bf P&\bf F1\\\midrule
Recent-2&-&-&-&-&31.7&42.2&36.2&10.3&15.6&12.4&5.0&6.7&5.7\\
Recent-3&-&-&-&-&31.7&42.2&36.2&16.9&17.0&17.0&0.0&0.0&0.0\\
Recent-4&-&-&-&-&30.0&40.9&34.6&18.4&14.2&16.0&0.0&0.0&0.0\\
Recent-5&-&-&-&-&28.3&39.5&33.0&16.9&10.7&13.1&0.0&0.0&0.0\\
Random&-&-&-&-&31.7&42.2&36.2&5.9&3.7&4.5&0.0&0.0&0.0\\\midrule
{\JOINT}&77.1&\bf 72.6&77.2&74.8&\bf 50.0&51.7&50.9&\bf 39.0&35.3&\bf 37.1&15.0&15.5&15.3\\
{\PRETRAIN}&\bf 77.9&72.4&\bf 78.0&\bf 75.1&45.0&\bf 71.1&\bf 55.1&30.2&\bf 46.1&36.4&\bf 16.7&\bf 26.3&\bf 20.4\\
\bottomrule
\end{tabular}}
\caption{\label{table:separate} 
Separate evaluation of our systems on the test set. X$_{split}$ are the scores for the split-antecedent anaphors.
}
\end{table*}

\begin{table}[t]
\centering
\resizebox{\columnwidth}!{
\begin{tabular}{lcccccc}
\toprule
&\multicolumn{3}{c}{\bf Imp$_{split}$ = 1}&\multicolumn{3}{c}{\bf Imp$_{split}$ = 10}\\
&\bf R&\bf P&\bf F1&\bf R&\bf P&\bf F1\\\midrule
Recent-2&70.5&66.9&68.7&61.5&61.3&61.4\\
Recent-3&70.5&66.9&68.7&61.6&61.1&61.4\\
Recent-4&70.6&66.9&68.7&61.8&61.1&61.5\\
Recent-5&70.5&66.9&68.7&61.5&61.2&61.3\\
Random&70.4&66.7&68.5&60.9&60.0&60.4\\
\midrule
Our model&\bf 70.8&\bf 67.2&\bf 69.0&\bf 63.8&\bf 64.4&\bf 64.1\\
\bottomrule
\end{tabular}}
\caption{\label{table:lea} 
{\LEA} evaluation on both single- and split-antecedent anaphors. Imp$_{split}$ indicates the split-antecedent importance. 
}
\end{table}

\section{Results and Discussions}
\subsection{Separate Evaluation on Single/Split-antecedent Anaphors}

We first evaluate our two proposed systems in the separate evaluation setting, in which we report separate scores for single-antecedent anaphors, non-referring expressions and split-antecedent anaphors. 
Showing  individual scores for different aspects provide  a  clear picture of the different models. 

\paragraph{Training settings}
In the \textbf{{\JOINT}} setting, the system is trained end-to-end with a weighted loss function. 
In the \textbf{{\PRETRAIN}} setting, we use the pretrained model provided by \newcite{yu-etal-2020-cluster}, and train only the split-antecedent part of the system. 

\paragraph{Baselines}
Like \cite{vala-etal-2016-antecedents,yu-etal-2020-plural}, we include 
baselines  based on heuristic rules or random selection. 
For all 
baselines, we use the same model as used by our {\PRETRAIN} approach to supply the candidate split-antecedent anaphors/singular clusters. 
The \textbf{anaphora recognition} baseline classifies as split-antecedent anaphors the discourse-new mentions  belonging to a small list of plural pronoun (e.g., \textit{they}, \textit{their}, \textit{them}, \textit{we}).\footnote{We also tried a random selection based approach, but such an approach only gets less than 5\% split-antecedent anaphors correctly.} 
The \textbf{recent-x baseline} chooses the $x$ closest singular clusters as  antecedents for these candidates. 
The  \textbf{random baseline}  assigns two to five antecedents randomly to each chose split-antecedent anaphors.

\paragraph{Results}
Table \ref{table:separate} shows the comparison between our two systems and the baselines. 
Since  plural pronouns are the most frequent split-antecedent anaphors, the simple heuristic gives a reasonably good F1 score of up to 36.2\% for  anaphora recognition. 
In term of the scores on full resolution, the baselines only achieved a maximum F1 of 17\% and 5.7\% when evaluated in the lenient and strict settings respectively. 
The low F1 scores indicate that split-antecedent anaphors are hard to resolve. 

When compared with the baselines, both of our approaches achieved much better scores for all three evaluations. 
Our models achieved substantial improvements over the baselines of up to 19\%, 19.9\% and 14.7\% for anaphora recognition, full resolution (lenient and strict) respectively.  
The model trained in a {\JOINT} setting achieves a better recall for both lenient evaluation and anaphora recognition, while the {\PRETRAIN} setting has much better precision. 
We expect this is  because the joint system could have an impact on candidate mentions/clusters, hence  potentially recover more antecedent-anaphora pairs. By contrast, the candidate mentions/clusters are fixed in the {\PRETRAIN} setting. Overall, the {\JOINT} model achieves a slightly better lenient F1 score but a lower strict F1 score, whereas the {\PRETRAIN} setting has a better overall performance when compared with the {\JOINT} model. The {\JOINT} system also has a lower {\CONLL} average F1 score and non-referring F1 score when compared with the system trained in a {\PRETRAIN} fashion. This indicates that jointly training is not helpful for the single-antecedent anaphors and non-referring expressions. Hence we use the {\PRETRAIN} approach for further experiments.

\subsection{Evaluating single and split antecedent anaphors jointly}
We then evaluate our models with the newly extended {\LEA} scores to show how split-antecedent anaphors could impact the results when evaluated together with single-antecedent anaphors. Table \ref{table:lea} shows the {\LEA} score comparison between our best model ({\PRETRAIN}) and the baselines. 
As only half of the test documents contain split-antecedent anaphors, we report the results on those test documents to give a clear picture on the evaluation.

We carried out two evaluations. 
The first setting is the traditional evaluation setting for coreference, in which split-antecedent anaphors are weighed equally as single antecedent anaphors (i.e., they are treated in {\LEA} as a single mention, Imp$_{split}$ = 1). 
We do not believe, however, that treating all anaphors equally is the most informative approach to evaluating coreference, for it is well-known that some anaphors are much easier to resolve than others \cite{barbu&mitkov:ACL2001,webster-et-al:TACL2018}. 
{\LEA} makes it possible to give more weight to anaphors that are harder to resolve.  
So in our second evaluation we give more importance to  split-antecedent anaphors (Imp$_{split}$ = 10) since they are much harder to resolve and also infrequent when compare to the single-antecedent anaphors. To have slightly higher importance for split-antecedents will give us a better view of their impact. 
The results in Table \ref{table:lea}  show that our best model achieved moderate improvements of  0.3\% - 0.5\% on the first {\LEA} score setting when compared with the baselines. 
This is mainly because the split-antecedent anaphors are less than 1\% of the mentions. 
But ss expected,  the improvements become more clear in the second evaluation setting, in which our model is 2.6\% - 3.7\% better than the baselines. 

\subsection{State-of-the-art Comparison}

To compare with the state-of-the-art system on {\ARRAU}, \cite{yu-etal-2020-plural},  we train our best setting ({\PRETRAIN})  as \newcite{yu-etal-2020-plural} did, i.e., assuming both gold mention and gold split-antecedent anaphors are provided. 
We first train the base model using gold mentions, then train the split-antecedent part of the system using gold mentions and gold split-antecedent anaphors. Since \newcite{yu-etal-2020-plural}'s system is evaluated on the full {\ARRAU} corpus and with a customised train/test split priorities the split-antecedent anaphors, we retrain their system using the same standard \ACRO{rst} split as used in our evaluation. 
We train their system with both  baseline and the best settings using a single auxiliary corpus (\textsc{single-coref}).\footnote{The best setting, that uses multi-auxiliary corpora, is more complex to train and only  moderately improves the results.} 
As shown in Table \ref{table:sota}, our best model achieved both better lenient and better strict accuracy than the \newcite{yu-etal-2020-plural} system, even though theirs  is a dedicated system concerned only with split-antecedent anaphora. The results suggest the pre-trained mention/cluster representations are suitable  for low-resource tasks that reply heavily on such representations.

\begin{table}[t]
\centering
\resizebox{\columnwidth}!{
\begin{tabular}{lcccc}
\toprule
&\multicolumn{3}{c}{\bf Lenient}&\bf Strict\\
&\bf R&\bf P&\bf F1&\bf Accuracy\\\midrule
Yu et al. Baseline&61.0&52.5&56.5&21.7\\
Yu et al. Best model&69.1&63.9&66.4&35.0\\\midrule
Our model&\bf 71.3&\bf 65.1&\bf 68.1&\bf 45.0\\
\bottomrule
\end{tabular}}
\caption{\label{table:sota} 
State-of-the-art comparison on the test set. 
}
\vspace{-20pt}
\end{table}

\subsection{Analysis}
In this section, we carry a qualitative analysis on the system outputs to find out the main courses of the performance gaps between the gold and predicted settings. 
We also report a more detailed comparison between our system and the \newcite{yu-etal-2020-plural} system to see if there is a systematic difference between the two systems on the gold settings.

\textbf{The Challenge of Using Predicted Setting}
The split-antecedent anaphora resolution task is more complex than its single-antecedent counterpart. 
The semantic relation between each individual antecedent and the anaphora is not identity, but element-of;
and the number of antecedents can also vary. 
The results on evaluations with gold mentions and gold split-antecedent anaphors provided are promising. However, when evaluated using  predicted mentions we have two main challenges:  anaphora recognition and noisy candidate mentions/clusters. 
For anaphora recognition, our best model ({\PRETRAIN}) only recalls 45\% of the anaphors. 
The performance of our anaphora recognition is affected by the predicted mentions, and further capped by the fact that we only attempt to classify as split-antecedent the mentions classed as  discourse-new by the base model. 
To assess the impact of these two factors, 
we computed the recall of  split-antecedent anaphors by  predicted mentions and 
discourse-new mentions. 
Virtually all split-antecedent anaphors are recalled among the predicted mentions--98.33\%--but only 65\% are recalled among the discourse-new mentions. 
This has a big impact on our results for split anaphora recognition, since  35\% of the anaphors are not accessible to our system. 
To understand the impact of 
this gap 
on the result, we supply to our system 
the 98.33\% of split-antecedent anaphors recognized as  predicted. We keep everything else-- predicted mentions, and clusters--unchanged. 
When run this way, our system achieves a lenient F1 score of 47.7\%, which is  11.3 p.p. better than the score (36.4\%) achieved using predicted anaphors, although   still 20.4\% lower than the model trained and evaluated with  gold mentions and gold split-antecedent anaphors (68.1\%). 
We suggest this additional difference is mainly a result of  noise in the predicted mentions and clusters. Overall, then, the noise in the predicted mentions and clusters contributed 2/3 of the score difference, while  problems with anaphora recognition are responsible for the rest.

\textbf{In Depth Comparison with \newcite{yu-etal-2020-plural}}. 
Next, we compared our model's  outputs in the gold setting 
with those of 
the best model of \newcite{yu-etal-2020-plural} in more detail. 
We split the test set in two different ways and compute the system performances on different categories. 
First, we follow \newcite{yu-etal-2020-plural} and split the split antecedent anaphors in the test set into two classes according to  the number of gold split-antecedents: 
one class includes the anaphors with two split-antecedents, 
whereas the second class includes the anaphors with  three or more split-antecedents (about 23\% of the total). 
Table \ref{table:m_anaphra} compares these two classes.
As we can see from the Table, with lenient evaluation the two systems work equally well for the anaphors with two split-antecedents,
but  our model is 8.5\% better for mentions with three or more split-antecedents. 
In terms of strict evaluation, our model outperforms the \cite{yu-etal-2020-plural} model by 8.7\% and 14.3\% for two classes respectively. 
Overall, the model presented here achieved substantial performance gains on anaphors with three or more split-antecedents.

We then split the data into two classes according to a different criterion: the part-of-speech of the anaphor. 
The first class consists of pronoun anaphors, such as ``they'' or ``their''.
The second class consists of all other split antecedent anaphors, such as  ``those companies'' or ``both''. 
As shown in Table \ref{table:they_anaphra}, the \cite{yu-etal-2020-plural} model achieves better scores for pronoun anaphors (mainly ``they'' and ``their''). However, our new model outperforms the old system with non-pronominal anaphors   by 5.4\% according to lenient F1, and doubled their strict accuracy. 

\begin{table}[t]
\centering
\resizebox{\columnwidth}!{
\begin{tabular}{lccccc}
\toprule
&&\multicolumn{2}{c}{\bf Yu et al.}&\multicolumn{2}{c}{\bf Our model}\\
&\bf Count&\bf Lenient&\bf Strict &\bf Lenient&\bf Strict \\
\midrule
2   &46     &71.9  &45.7    &70.9   &54.4\\
3+  &14     &52.5  &0.0     &61.0   &14.3\\
\bottomrule
\end{tabular}}
\caption{\label{table:m_anaphra} Scores for anaphors with different number of antecedents.}
\end{table}

\begin{table}[t]
\centering
\resizebox{\columnwidth}!{
\begin{tabular}{lccccc}
\toprule
&&\multicolumn{2}{c}{\bf Yu et al.}&\multicolumn{2}{c}{\bf Our model}\\
&\bf Count&\bf Lenient&\bf Strict &\bf Lenient&\bf Strict \\
\midrule
PRP   &24     &82.4  &58.3     &76.4   &54.2\\
Other  &36    &57.5  &19.4     &62.9   &38.9\\
\bottomrule
\end{tabular}}
\caption{\label{table:they_anaphra} Scores for pronoun and other anaphors.}
\end{table}

\section{Conclusions}

In this paper, we introduced a neural system performing both single and split-antecedent anaphora resolution, and evaluated the system in a more realistic setting than previous work. 
We extended the state-of-the-art coreference system on {\ARRAU} to also resolve  split-antecedent anaphors. 
The proposed system achieves much better results on split-antecedent anaphors when compared with the baselines using heuristic and random selection when using the predicted mentions/clusters. 
Our system also achieves better results  than  the previous state-of-the-art system on {\ARRAU} \cite{yu-etal-2020-plural}, which only attempted single-antecedent anaphora resolution from gold mentions, 
when evaluated on the same task. 

In addition, we also proposed an extension of the  {\LEA} coreference evaluation metric to evaluate both single and split-antecedent anaphors in a single metric.

\section*{Acknowledgements}
This research was supported in part by the DALI project, ERC Grant 695662.


\bibliography{acl}
\bibliographystyle{acl_natbib}


\end{document}